\documentclass[sigconf]{acmart}

\usepackage{tabularx}
\usepackage{booktabs} 
\usepackage{array}
\usepackage{subfigure}
\usepackage{listings}
\usepackage{xcolor}
\usepackage{mathrsfs}
\usepackage{fancyhdr}

\usepackage[norndcorners,customcolors]{hf-tikz}
\hfsetbordercolor{yellow}
\hfsetfillcolor{yellow}
\definecolor{mygreen}{rgb}{0,0.6,0}
\definecolor{mygray}{rgb}{0.5,0.5,0.5}
\definecolor{mymauve}{rgb}{0.58,0,0.82}

\setcopyright{none}

\begin{document}
\title{Synetgy: Algorithm-hardware Co-design for ConvNet Accelerators on Embedded FPGAs}

\author{Yifan Yang$^{1,2,*}$, Qijing Huang$^1$, Bichen Wu$^1$, Tianjun Zhang$^1$, Liang Ma$^3$, Giulio Gambardella$^4$, Michaela Blott$^4$, Luciano Lavagno$^3$, Kees Vissers$^4$, John Wawrzynek$^1$, Kurt Keutzer$^1$}
\affiliation{$^1$UC Berkeley; $^2$Tsinghua University; $^3$Politecnico di Torino; $^4$Xilinx Research Labs}  
\email{{yifan-yang, qijing.huang, bichen, tianjunz, johnw, keutzer}@berkeley.edu;  } 
\email{{luciano.lavagno, liang-ma}@polito.it;{giuliog, mblott, keesv}@xilinx.com }

\thanks{*Work done while interning at UC Berkeley.}

\begin{abstract}
Using FPGAs to accelerate ConvNets has attracted significant attention in recent years. However, FPGA accelerator design has not leveraged the latest progress of ConvNets. As a result, the key application characteristics such as frames-per-second (FPS) are ignored in favor of simply counting GOPs, and results on accuracy, which is critical to application success, are often not even reported. In this work, we adopt an algorithm-hardware co-design approach to develop a ConvNet accelerator called Synetgy and a novel ConvNet model called DiracDeltaNet$^{\dagger}$. Both the accelerator and ConvNet are tailored to FPGA requirements. DiracDeltaNet, as the name suggests, is a ConvNet with only $1\times 1$ convolutions while spatial convolutions are replaced by more efficient shift operations. DiracDeltaNet achieves competitive accuracy on ImageNet (88.7\% top-5), but with 42$\times$ fewer parameters and 48$\times$ fewer OPs than VGG16. We further quantize DiracDeltaNet's weights to 4-bit and activations to 4-bits, with less than 1\% accuracy loss. These quantizations exploit well the nature of FPGA hardware. In short, DiracDeltaNet's small model size, low computational OP count, low precision and simplified operators allow us to co-design a highly customized computing unit for an FPGA. We implement the computing units for DiracDeltaNet on an Ultra96 SoC system through high-level synthesis. Our accelerator's final top-5 accuracy of 88.1\% on ImageNet, is higher than all the previously reported embedded FPGA accelerators. In addition, the accelerator reaches an inference speed of 66.3 FPS on the ImageNet classification task, surpassing prior works with similar accuracy by at least 11.6$\times$.
\end{abstract}

\thanks{$\dagger$ Source code and pre-trained model are available at \url{https://github.com/Yang-YiFan/DiracDeltaNet}.}
%
%

\settopmatter{printacmref=false} 
\renewcommand\footnotetextcopyrightpermission[1]{}
\pagestyle{plain} 

\maketitle

\section{Introduction}
ConvNets power state-of-the-art solutions on a wide range of computer vision tasks. However, the high computational complexity of ConvNets hinders their deployment on embedded and mobile devices, where computational resources are limited. Using FPGAs to accelerate ConvNets has attracted significant research attention in recent years. FPGAs excel at low-precision computation, and their adaptability to new algorithms lends themselves to supporting rapidly changing ConvNet models.

Despite recent efforts to use FPGAs to accelerate ConvNets, as \cite{kwon2018co} points out, there still exists a wide gap between accelerator architecture design and ConvNet model design. The computer vision community has been primarily focusing on improving the accuracy of ConvNets on target benchmarks with only secondary attention to the computational cost of ConvNets. As a consequence, recent ConvNets have been trending toward more layers \cite{he2016identity}, more complex structures \cite{huang2017densely, zoph2017learning}, and more complicated operations \cite{yu2015multi}. 

On the other hand, FPGA accelerator design has not leveraged the latest progress of ConvNets. Many FPGA designs still focus on networks trained on CIFAR10 \cite{krizhevsky2009learning}, a small dataset consisting of 32x32 thumbnail images. Such dataset is usually used for experimental purposes and is too small to have practical value. More recent designs aim to accelerate inefficient ConvNets such as AlexNet \cite{krizhevsky2012imagenet} or VGG16 \cite{simonyan2014very}, both of which have fallen out of use in state-of-the-art computer vision applications. In addition, we observe that in many previous designs, key application characteristics such as frames-per-second (FPS) are ignored in favor of simply counting GOPs, and accuracy, which is critical to applications, is often not even reported. 

Specifically, we see a gap between ConvNet architectures and accelerator design in the following areas: 

\textbf{Inefficient ConvNet models}: Many FPGA accelerators still target older, inefficient models such as AlexNet and VGG16, which require orders-of-magnitude greater storage and computational resources than newer, efficient models that achieve the same accuracy. With an inefficient model, an accelerator with high throughput in terms of GOPs can actually have low inference speed in terms of FPS, where FPS is the more essential metric of efficiency. To achieve AlexNet-level accuracy, SqueezeNet \cite{iandola2016squeezenet} is 50x smaller than AlexNet; SqueezeNext \cite{gholami2018squeezenext} is 112x smaller; ShiftNet-C \cite{wu2017shift}, with 1.6\% higher accuracy, is 77x smaller. However, not many designs target those efficient models. Additionally, techniques for accelerating older models may not generalize to newer ConvNets. 

\textbf{ConvNet structures}: Most ConvNets are structured solely for better accuracy. Some ConvNets are structured for optimal GPU efficiency, but few, if any, are designed for optimal FPGA efficiency. For example, the commonly used additive skip connection \cite{he2016deep} alleviates the difficulty of training deep ConvNets and significantly boosts accuracy. Despite its mathematical simplicity, the additive skip connection is difficult to efficiently implement on FPGAs. Additive skip connections involve adding the output data from a previous layer to the current layer, which requires either using on-chip memory to buffer the previous layer's output or fetching the output from off-chip memory. Both options are inefficient on FPGAs.

\textbf{ConvNet operators}: ConvNet models contain many different types of operators. Commonly used operators include 1$\times$1, 3$\times$3, 5$\times$5 convolutions, 3$\times$3 max-pooling, etc. More recent models also contain depth-wise, group, dilated, and factorized convolutions. Not all of these operators can be efficiently implemented on FPGAs. If a ConvNet contains many different types of operators, one must either allocate more dedicated compute units or make the compute unit more general. Either solution can potentially lead to high resource requirement, limited parallelism, and more complicated control flow. Also, hardware development will require more engineering effort.

\textbf{Quantization}: ConvNet quantization has been widely used to convert weights and activations from floating point to low-precision numbers to reduce the computational cost. However, many of the previous methods are not practically useful for FPGAs due to the following problems: 1) Quantization can lead to serious accuracy loss, especially if the network is quantized to low precision numbers (less than 4 bits). Accuracy is vital for many computer vision applications. Unfortunately, carefully reporting accuracy has not been the norm in the FPGA community. 2) Many of the previously presented quantization methods are only effective on large ConvNet models such as VGG16, AlexNet, ResNet, etc. Since those models are known to be redundant, quantizing those to low-precision is much easier. We are not aware of any previous work tested on efficient models such as MobileNet or ShuffleNet. 
3) Many methods do not quantize weights and activations directly to fixed point numbers. Usually, quantized weights and activations are represented by fixed-point numbers multiplied by some shared floating point coefficients. Such representation requires more complicated computation than purely fixed-point operations, and are therefore more expensive.

In this work, we adopt an algorithm-hardware co-design approach to develop a ConvNet accelerator called Synetgy and a novel ConvNet model called DiracDeltaNet. Both the accelerator and the ConvNet are tailored to FPGAs and are optimized for ImageNet classification accuracy and inference speed (in terms of FPS). Our co-design approach produces a novel ConvNet architecture DiracDeltaNet that is based on ShuffleNetV2 \cite{ma2018shufflenet}, one of the state-of-the-art efficient models with small model size, low FLOP counts, hardware friendly skip connections, and competitive accuracy. We optimize the network by replacing all 3$\times$3 convolutions with shift operations \cite{wu2017shift} and 1$\times$1 convolution, enabling us to implement a compute unit customized for 1$\times$1 convolutions for better efficiency. The name ``DiracDeltaNet'' comes from the fact that the network only convolves input feature maps with 1$\times$1 kernels. Such kernel functions can be seen as discrete 2D Dirac Delta functions. We further quantize the network to 4-bit weights and 4-bit activations, exploiting the strengths of FPGAs, with only a less than 1\% accuracy drop.  
In short, DiracDeltaNet's small model size, low operation count, low precision and simplified operators allow us to co-design a highly customized and efficient FPGA accelerator. Furthermore, the implementation only took two people working for one month using High-Level Synthesis (HLS).

We trained DiracDeltaNet on ImageNet, implemented it on our accelerator architecture, Synetgy, and deployed on a low-cost FPGA board (Ultra96). Our inference speed reaches 66.3 FPS, surpassing previous works with similar accuracy by at least 11.6x. The DiracDeltaNet on our accelerator architecture also achieves 88.1\% top-5 classification accuracy -- the highest among all the previously reported embedded FPGA accelerators.

\section{Background}
\subsection{Efficient ConvNet Models}
For the task of image classification, improving accuracy on the ImageNet \cite{deng2009imagenet} dataset has been the primary focus of the computer vision community. For applications that are sensitive to accuracy, even a 1\% improvement in accuracy on ImageNet is worth doubling or tripling model complexity. As a concrete example, ResNet152 \cite{he2016deep} achieves 1.36\% higher ImageNet accuracy than ResNet50 at the cost of 3x more layers. In recent years, efficient ConvNet models have begun to receive more research attention. SqueezeNet \cite{iandola2016squeezenet} is one of the early models focusing on reducing the parameter size. While SqueezeNet is designed for image classification, later models, including SqueezeDet \cite{wu2017squeezedet} and SqueezeSeg \cite{wu2017squeezeseg, wu2018squeezesegv2}, extend the scope to object detection and point-cloud segmentation. More recent models such as MobileNet \cite{howard2017mobilenets, sandler2018mobilenetv2} and ShuffleNet \cite{zhang1707shufflenet, ma2018shufflenet} further reduce model complexity. However, without a target computing platform in mind, most models designed for ``efficiency'' can only target intermediate proxies to efficiency, such as parameter size or FLOP count, instead of focusing on more salient efficiency metrics, such as speed and energy. Recent works also try to bring in hardware insight to improve the actual efficiency. SqueezeNext\cite{gholami2018squeezenext} uses a hardware simulator to adjust the macro-architecture of the network for better efficiency. ShiftNet\cite{wu2017shift} proposes a hardware-friendly shift operator to replace expensive spatial convolutions. AddressNet\cite{zhong2018rejecteccv} designed three shift-based primitives to accelerate GPU inference.

\subsection{ConvNet Quantization}
ConvNet quantization aims to convert full-precision weights and activations of a network to low-precision representations to reduce the computation and storage cost. Early works \cite{han2015deep, zhu2016trained} mainly focus on quantizing weights while still using full-precision activations. Later works \cite{rastegari2016xnor,zhou2016dorefa,choi2018pact,Zhuang2017progressive} quantize both weights and activations. Many previous works \cite{zhu2016trained,rastegari2016xnor,zhou2016dorefa} see serious accuracy loss if the network is quantized to low precisions. Normally, an accuracy loss of more than 1\% is already considered significant. Also, in many works \cite{zhu2016trained,choi2018pact}, quantized weights or activations are represented by low-precision numbers multiplied with some floating point coefficients. This can bring several challenges to hardware implementation. Last, but not least, most of the previous works report quantization results on inefficient models such as VGG, AlexNet, and ResNet. Given that those models are redundant, quantizing them to lower precisions is much easier. We have not yet seen any work which successfully applies quantization to efficient models.

\subsection{Hardware Designs}
Most existing ConvNet hardware research has focused on improving the performance of either standalone $3 \times 3$ convolution layers or a full-fledged, large ConvNet on large FPGA devices.
\cite{zhang2015optimizing} quantitatively studies the computation throughput and memory bandwidth requirement for ConvNets. 
\cite{zhang2017improving, ma2017optimizing} present their own optimization for ConvNets based on analytical performance models. They achieve high throughput on VGG16 using their proposed design methodology with OpenCL.
\cite{zhang2017frequency} designs convolution in frequency domain to reduce the compute intensity of the ConvNet. They demonstrate good power performance results on VGG16, AlexNet, and GoogLeNet.
\cite{nurvitadhi2017can} implements a ternary neural network on high-end Intel FPGAs and achieves higher performance/Watt than Titan X GPU.  
Most of the works mentioned above and others \cite{li2016high, aydonat2017opencl, wei2017automated}, target inefficient ConvNets on middle to high-end FPGA devices. 
For compact ConvNets, 
\cite{umuroglu2017finn} demonstrates a binary neural network(BNN) FPGA design that performs CIFAR10 classification at 21906 frames per second(FPS) with 283 $\mu$s latency on Xilinx ZC706 device. The BNN reports an accuracy of 80.1\%.  \cite{zhao2017accelerating, nakahara2017fully} run the BNN on a smaller device ZC7020. Although all three works achieve promising frame rates, they have not implemented larger neural networks for the ImageNet classification. It should be noted that classification on CIFAR10 dataset is orders of magnitude simpler than ImageNet, since CIFAR10 contains 100x fewer classes, 26x fewer images, and 49x fewer pixels in each image. Networks trained on CIFAR10 dataset also have way smaller complexity compared to those trained on ImageNet. In comparison, networks for ImageNet classification are closer to real-world applicability. \cite{qiu2016going} first attempted to deploy VGG16 for ImageNet classification on embedded device zc7020 and achieved a frame rate of 4.45 fps. Later \cite{guo2017software} improved the frame rate to 5.7 fps. However, their frame rate was relatively low for real-time image classification tasks. \cite{blott2018finnr, jiao2017accelerating, qiu2016going} have achieved high frame rate on smaller devices, however, the accuracy of their network is not on par with \cite{guo2017software} for ImageNet classification.

\section{ConvNet Design}
\label{NNDesign}
We discuss the ConvNet design in this section. 
The design of our ConvNet incorporates the feedback from both the computer vision applications and hardware accelerator design. Specifically, an ideal ConvNet model for embedded FPGA acceleration should satisfy the following aspects: 
1) The network should not contain too many parameters or FLOPs but should still maintain a competitive accuracy. 
2) The network structure should be hardware friendly to allow efficient scheduling. 
3) The network's operation set should be simplified for efficient FPGA implementation. 
4) The network's weights and activations should be quantized to low-precision fixed-point numbers without much accuracy loss. 

\subsection{ShuffleNetV2}
We select ShuffleNetV2-1.0x \cite{ma2018shufflenet} as our starting point. ShuffleNetV2 is one of the state-of-the-art efficient models. It has a top-1 accuracy of 69.4\% on ImageNet (2\% lower than VGG16), but contains only 2.3M parameters (60x smaller than VGG16) and 146M FLOPs (109x smaller than VGG16).

\begin{figure}[!t]
\begin{center}
\subfigure[ShuffleNetV2 blocks \cite{ma2018shufflenet}. ]{
\label{fig:shufflenet-blocks}
\includegraphics[width=.8\linewidth]{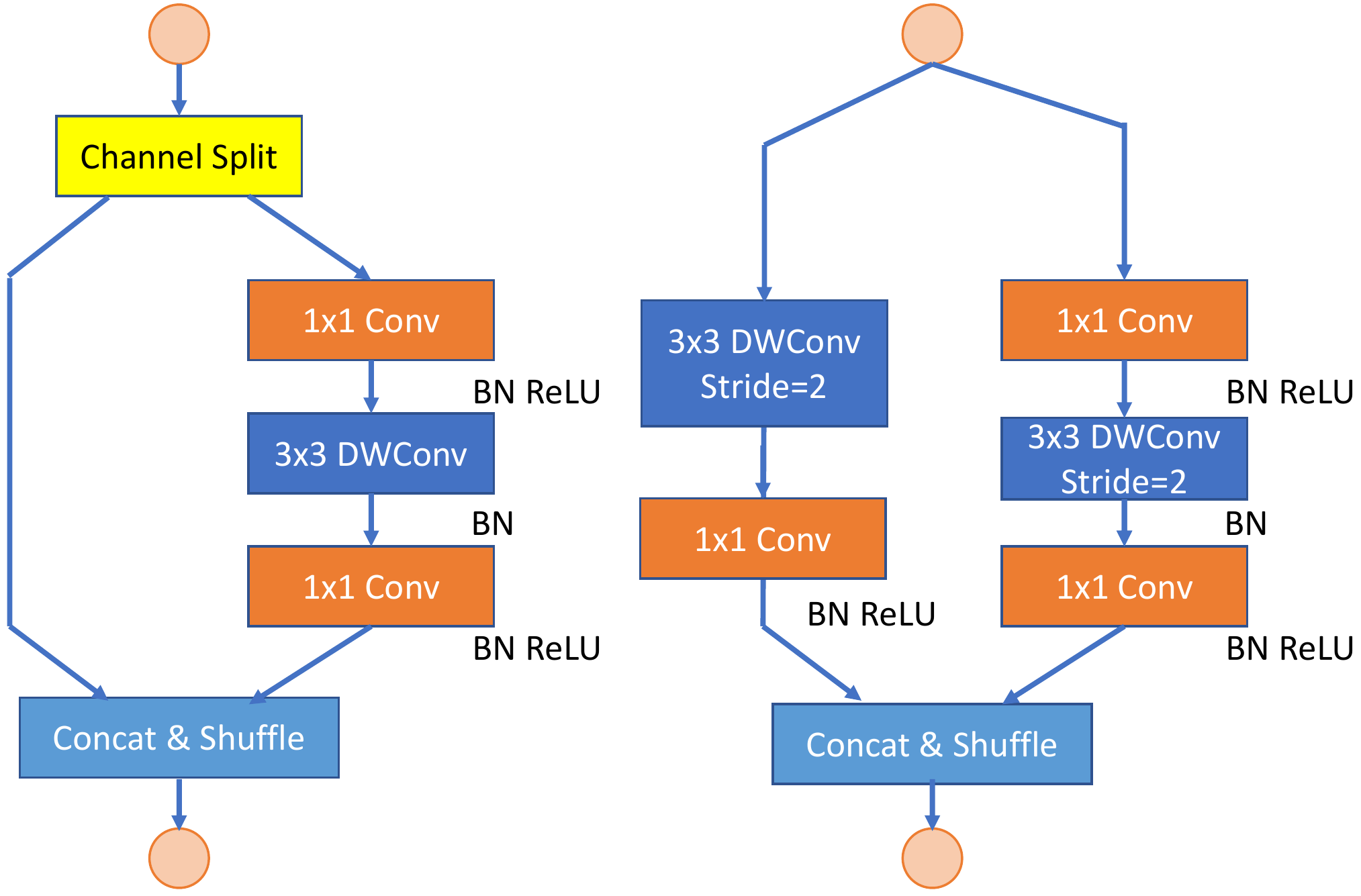}
}
\subfigure[Our modified DiracDeltaNet blocks. We replace depth-wise convolutions with shift operations. In the downsampling blocks, we use stride-2 max-pooling and shift operations to replace stride-2 depth-wise convolutions. We also double the filter number of the 1st 1$\times$1 convolution on the non-skip branch in each module.]{
\label{fig:dirac-block}
\includegraphics[width=.8\linewidth]{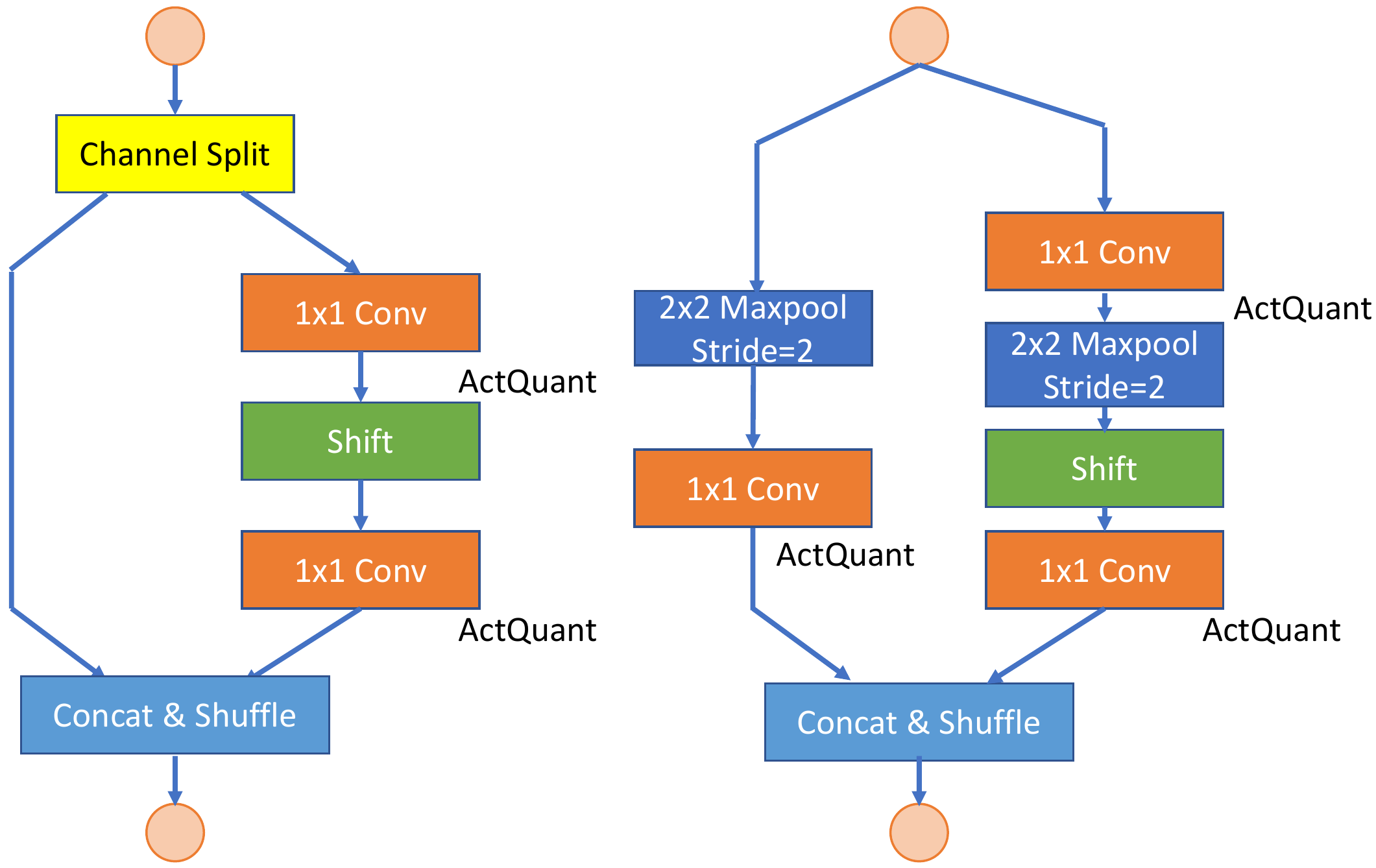}
}
\vspace{-10pt}
\caption{ShuffleNetV2 blocks vs. DiracDeltaNet blocks}
\label{fig:netblocks}
\end{center}
\vspace{-13pt}
\end{figure}

The block-level structure of ShuffleNetV2 is illustrated in Fig. \ref{fig:shufflenet-blocks}. The input feature map of the block is first split into two parts along the channel dimension. The first branch of the network does nothing to the input data and directly feeds the input to the output. The second branch performs a series of 1$\times$1 convolutions, 3$\times$3 depth-wise convolutions and another 1$\times$1 convolution operations on the input. Outputs of two branches are then concatenated along the channel dimension. Channel shuffle \cite{zhang1707shufflenet} is then applied to exchange information between branches. In down-sampling blocks, depth-wise 3$\times$3 convolutions with a stride of 2 are applied to both branches of the block to reduce the spatial resolution. 1$\times$1 convolutions are used to double the channel size of input feature maps. These blocks are cascaded to build a deep ConvNet. We refer readers to \cite{ma2018shufflenet} for the macro-structure description of the ShuffleNetV2.

\begin{figure}[!t]
\begin{center}
\centering \includegraphics[width=0.6\linewidth]{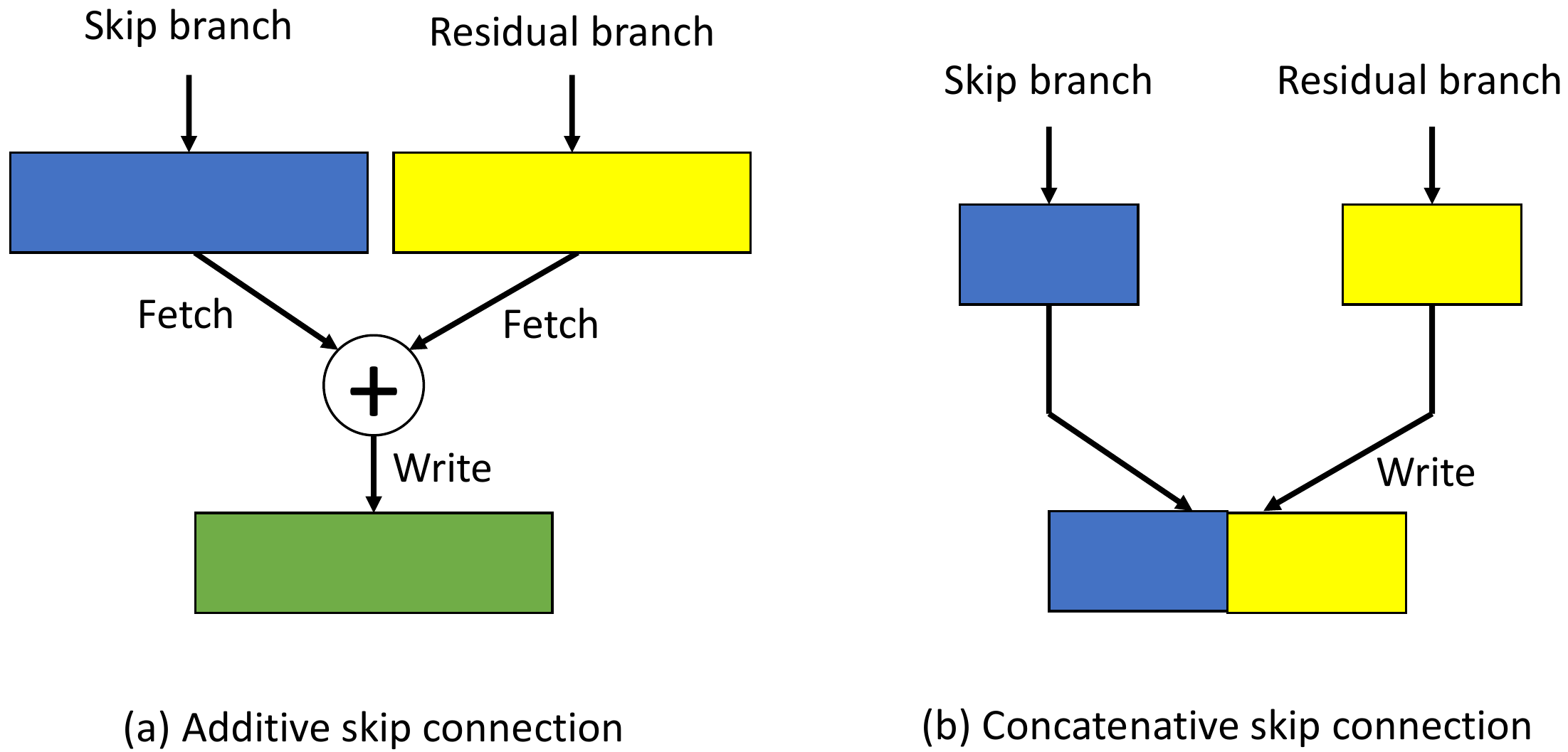}
\caption{Additive Skip Connections vs. Concatenative Skip Connections. Rectangles represent data tensors.}
\label{fig:add-vs-concat}
\end{center}
\vspace{-13pt}
\end{figure}

We select ShuffleNetV2-1.0x not only because of its small model size and low FLOP count but also because it uses concatenative skip connections instead of additive skip connections. Additive skip connections, as illustrated in Fig. \ref{fig:add-vs-concat}(a), were first proposed in \cite{he2016deep}. It effectively alleviates the difficulty of training deep neural networks and therefore improves accuracy. It is widely used in many ConvNet designs. However, additive skip connections are not efficient on FPGAs. As illustrated in Fig. \ref{fig:add-vs-concat}(a), both the skip and the residual branches' data need to be fetched on-chip to conduct the addition. Though addition does not cost too much computation, the data movement is expensive. Concatenative skip connections, as illustrated in Fig. \ref{fig:add-vs-concat}(b), were first proposed in \cite{huang2017densely}. It has a similar positive impact to the network training and accuracy. With concatenative skip connections, data from skip branch is already in off-chip DRAMs. So we can concatenate the two branches simply by writing the residual branch data next to the skip branch data. This avoids the extra memory access in additive skip connections and alleviates the memory bandwidth pressure.

\subsection{DiracDeltaNet}
\label{DiracDeltaNet}
Based on ShuffleNetV2, we build DiracDeltaNet through the following modifications: 1) we replace all the 3$\times$3 convolutions with shift and 1$\times$1 convolutions; 2) we reduce the kernel size of max-pooling from 3$\times$3 to 2$\times$2; 3) we modify the order of channel shuffle.

\begin{figure}[!t]
\begin{center}
\centering \includegraphics[width=1.0\linewidth]{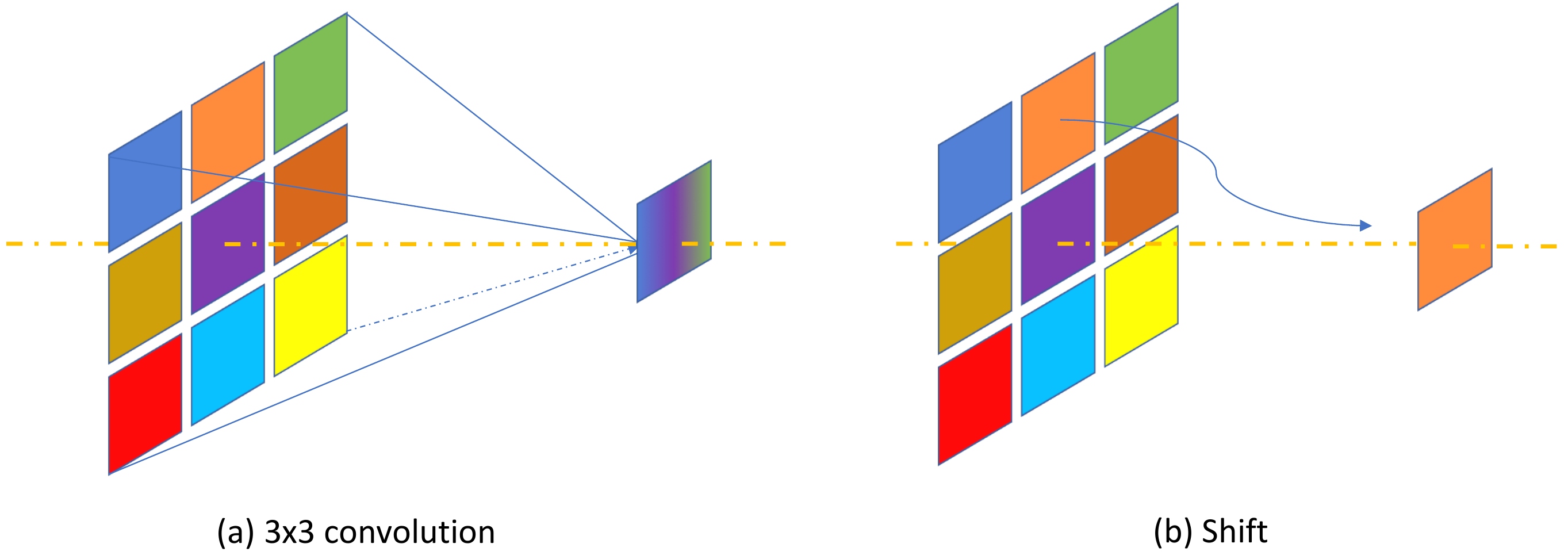}
1\caption{3$\times$3 Convolution vs. Shift. In 3$\times$3 convolutions, pixels in a 3$\times$3 region are aggregated to compute one output pixel at the center position. In the shift operation, a neighboring pixel is directly copied to the center position.}
\label{fig:shift}
\end{center}
\end{figure}

\begin{figure}[!t]
\begin{center}
\centering \includegraphics[width=0.75\linewidth]{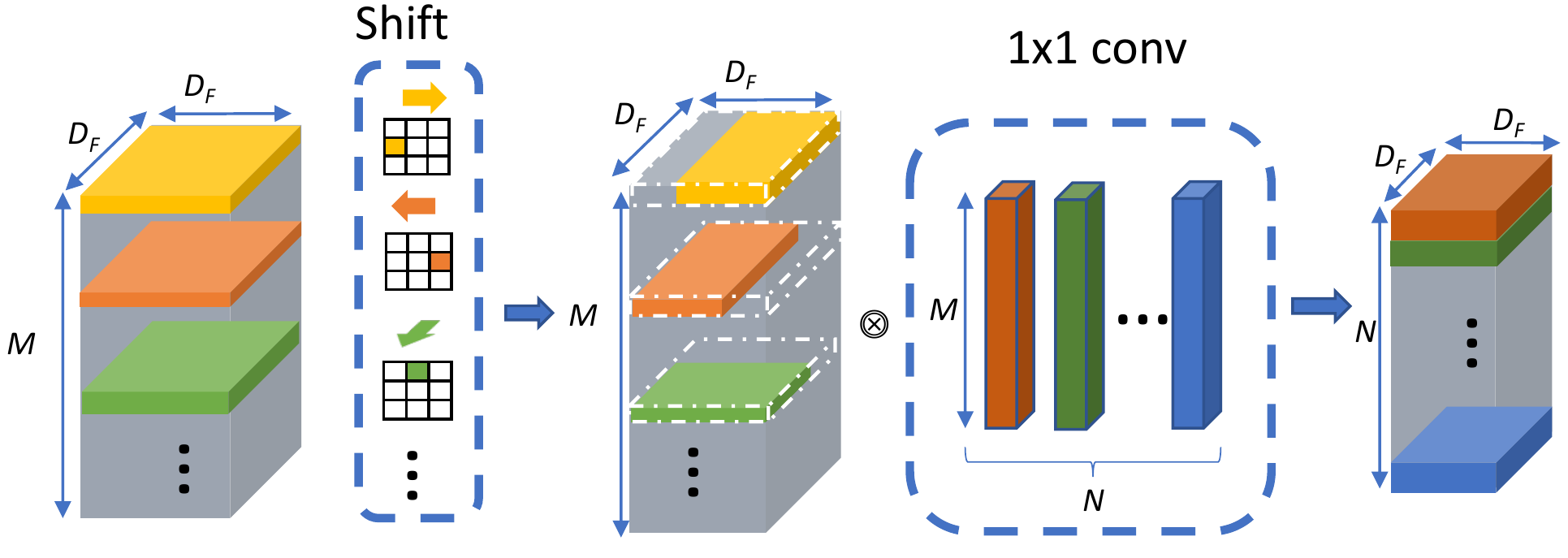}
\caption{Using shift and 1$\times$1 convolutions to replace 3$\times$3 convolutions. This figure is from \cite{wu2017shift}.}
\label{fig:shift-layer}
\end{center}
\end{figure}

We replace all of the 3$\times$3 convolutions and 3$\times$3 depth-wise convolutions with shift operations and 1$\times$1 convolutions. The motivation is that smaller convolution kernel sizes require less reuse of the feature map, resulting in simpler data movement schedule, control flow, and timing constraint. As pointed out by \cite{wu2017shift}, ConvNets rely on spatial convolutions (3$\times$3 convolutions and 3$\times$3 depth-wise convolutions) to aggregate spatial information from neighboring pixels to the center position. However, spatial convolutions can be replaced by a more efficient operator called shift. The shift operator aggregates spatial information by copying nearby pixels directly to the center position. This is equivalent to shifting one channel of feature map towards a certain direction. When we shift different channels in different directions, the output feature map's channel will encode all the spatial information. A comparison between 3$\times$3 convolution and shift is illustrated in Fig. \ref{fig:shift}. A module containing a shift and 1$\times$1 convolution is illustrated in Fig. \ref{fig:shift-layer}.

For 3$\times$3 depth-wise convolutions, we directly replace them with shift operations, as shown in Fig. \ref{fig:dirac-block}. This direct replacement can lead to some accuracy loss. To mitigate this, we double the output filter number of the first 1$\times$1 convolution on the non-skip branch from Fig. \ref{fig:dirac-block}. Nominally, doubling the output channel size increases both FLOP count and parameter size by a factor of 2. However, getting rid of 3$\times$3 convolutions allows us to design a computing unit customized for 1$\times$1 convolutions with higher execution efficiency than a comparable unit for 3$\times$3 depth-wise convolutions. In the downsample block, we directly replace the strided 3$\times$3 depth-wise convolutions with a stride-2 2$\times$2 max-pooling. Unlike \cite{wu2017shift}, our shift operation only uses 4 cardinal directions (up, down, left, right) in addition to the identity mapping (no-shift). This simplifies our hardware implementation of the shift operation without hurting accuracy. 

The first stage of ShuffleNetV2 consists of a 3$\times$3 convolution with a stride of 2 and filter number of 24. It is then followed by a 3$\times$3 max-pooling with a stride of 2. We replace these two layers to a module consisting of a series of 1$\times$1 convolution, 2$\times$2 max-pooling, and shift operations, as shown in Table \ref{tab:dirac-macro}. Compared with the original 3$\times$3 convolutions, our proposed module has more parameters (2144 vs 648) and FLOPs (30.5M vs 8.1M). But the implementation and execution cost of the proposed first stage is negligible compared to a 3$\times$3 convolution layer. After training the network, we find that this module gives near equal accuracy than the original 3$\times$3 convolution module. With our new module, we can eliminate the remaining 3$\times$3 convolutions from our network, enabling us to allocate more computational resources to 1$\times$1 convolutions, and thereby increasing parallelism and throughput. 

In addition to replacing all 3$\times$3 convolutions, we also reduce the max-pooling kernel size from 3$\times$3 to 2$\times$2. By using the same pooling kernel size as the stride, we eliminate the need to buffer extra data on the pooling kernel boundaries, thereby achieving better efficiency. Our experiments also show that reducing the max-pooling kernel size does not impact accuracy.

\begin{figure}[!t]
\begin{center}
\centering 
\includegraphics[width=1.0\linewidth]{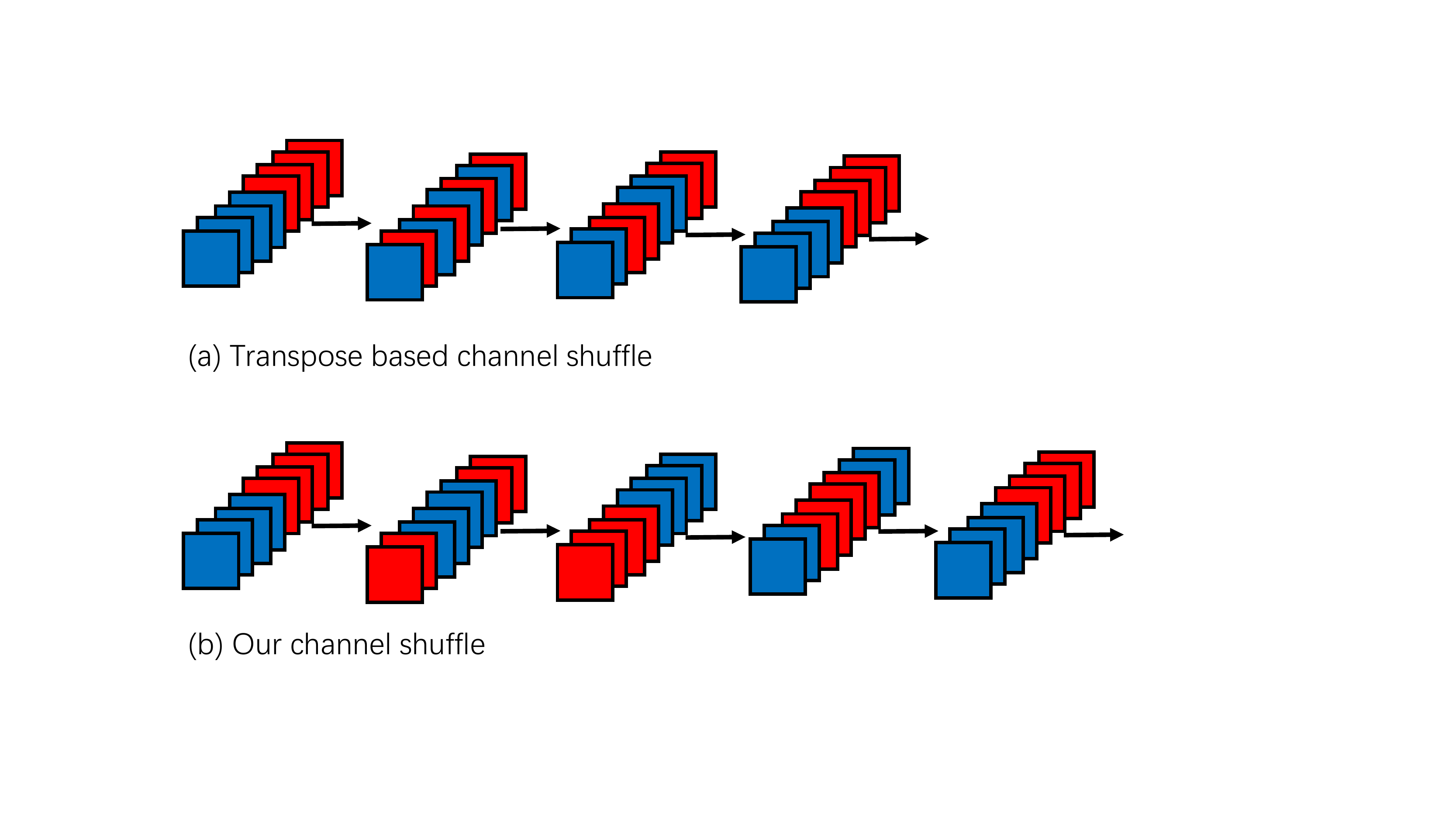}
\vspace{-17pt}
\caption{Transpose Based Shuffle (ShuffleNetV2) vs. Our HW Efficient Shuffle (DiracDeltaNet)}
\label{fig:shuffle}
\end{center}
\vspace{-13pt}
\end{figure}

We also modify the channel shuffle's order to make it more hardware efficient. ShuffleNetV2 uses transpose operation to mix channels from two branches. This is illustrated in Fig. \ref{fig:shuffle}(a), where blue and red rectangles represent channels from different branches. The transpose based shuffling is not hardware friendly since it breaks the contiguous data layout. Performing channel shuffle in this manner will require multiple passes of memory read and write. We propose a more efficient channel shuffle showed in Fig. \ref{fig:shuffle}(b). We perform a circular shift to the feature map along the channel dimension. We can have the same number of channels exchanged between two branches while preserving the contiguity of the feature map and minimizing the memory accesses.

We name the modified ShuffleNetV2-1.0x model as DiracDeltaNet. The name comes from the fact that our network only contains 1$\times$1 convolutions. With a kernel size of 1, the kernel functions can be seen as discrete 2D Dirac Delta functions. DiracDeltaNet's macro-structure is summarized in Table \ref{tab:dirac-macro}. Stage 2,3,4 consist of chained DiracDeltaNet blocks depicted in Fig. \ref{fig:netblocks} with different feature map size, channel size and stride. We adopt the training recipe and hyperparameters described in \cite{ma2018shufflenet}. We train DiracDeltaNet for 90 epoch with linear learning rate decay, the initial learning rate of 0.5, 1024 batch size and 4e-5 weight decay. A comparison between ShuffleNetV2-1.0x and our DiracDeltaNet is summarized in Table \ref{tab:net-compare}.

\begin{table}[]
\caption{Macro-structure of DiracDeltaNet}
\vspace{-5pt}
\begin{tabular}{c|ccccc}
\hline
Layer                                                                                     & \begin{tabular}[c]{@{}c@{}}Output\\ size\end{tabular}                   & \begin{tabular}[c]{@{}c@{}}Kernel\\ size\end{tabular}       & Stride                                                        & \#Repeat                                                      & \begin{tabular}[c]{@{}c@{}}Output \\ channel\end{tabular} \\ \hline
Image                                                                                     & 224                                                                     &                                                             &                                                               &                                                               & 3                                                         \\ \hline
\begin{tabular}[c]{@{}c@{}}Conv1\\ Maxpool\\ shift\\ Conv2\\ Maxpool\\ shift\end{tabular} & \begin{tabular}[c]{@{}c@{}}224\\ 112\\ 112\\ 112\\ 56\\ 56\end{tabular} & \begin{tabular}[c]{@{}c@{}}1\\ 2\\3\\1\\ 2\\ 3\end{tabular} & \begin{tabular}[c]{@{}c@{}}1\\ 2\\ 1\\ 1\\ 2\\ 1\end{tabular} & \begin{tabular}[c]{@{}c@{}}1\\ 1\\ 1\\ 1\\ 1\\ 1\end{tabular} & \begin{tabular}[c]{@{}c@{}}32\\ \\ \\ 64\end{tabular}     \\ \hline
Stage 2                                                                                    & \begin{tabular}[c]{@{}c@{}}28\\ 28\end{tabular}                         &                                                             & \begin{tabular}[c]{@{}c@{}}2\\ 1\end{tabular}                 & \begin{tabular}[c]{@{}c@{}}1\\ 3\end{tabular}                 & 128                                                       \\ \hline
Stage 3                                                                                   & \begin{tabular}[c]{@{}c@{}}14\\ 14\end{tabular}                         &                                                             & \begin{tabular}[c]{@{}c@{}}2\\ 1\end{tabular}                 & \begin{tabular}[c]{@{}c@{}}1\\ 7\end{tabular}                 & 256                                                       \\ \hline
Stage 4                                                                                   & \begin{tabular}[c]{@{}c@{}}7\\ 7\end{tabular}                           &                                                             & \begin{tabular}[c]{@{}c@{}}2\\ 1\end{tabular}                 & \begin{tabular}[c]{@{}c@{}}1\\ 3\end{tabular}                 & 512                                                       \\ \hline
Conv5   & 7    & 1   & 1     & 1    & 1024    \\ \hline
GlobalPool   & 1   & 7   &   &  1    & 1024 \\ \hline
FC     &   &    &   & 1   & 1000   \\ \hline
\end{tabular}
\label{tab:dirac-macro}
\vspace{-5pt}
\end{table}

\begin{table}[]
\caption{ShuffleNetV2-1.0x vs. DiracDeltaNet}
\vspace{-5pt}
\begin{tabular}{c|cccc}
\hline
                & MACs & \#Params & Top-1 acc & Top-5 acc \\ \hline
ShuffleNetV2-1.0x    & 146M  &  2.3M   & 69.4\%       &   -        \\ \hline
DiracDeltaNet & 330M  &  3.3M    & 68.9\%     &   88.7\%     \\ \hline
\end{tabular}
\label{tab:net-compare}
\vspace{-10pt}
\end{table}

\subsection{ConvNet Quantization}
\label{Quantization}

To further reduce the cost of DiracDeltaNet, we apply quantization to convert floating point weights and activations to low-precision integer values. For network weights, we follow DoReFa-Net \cite{zhou2016dorefa} to quantize full-precision weights as
\begin{equation}
\label{eqn:weight_quant}
w_k = 2Q_k(\frac{\tanh (w)}{2\text{max}(|\tanh (w)|)} + 0.5)-1.
\end{equation}
Here, $w$ denotes the latent full-precision weight of the convolution kernel. $Q_k(\cdot)$ is a function that quantizes its input in the range of $[0, 1]$ to its nearest neighbor in $\{\frac{i}{2^k-1}| i=0,\cdots 2^{k-1}\}$.  

We follow PACT \cite{choi2018pact} to quantize each layer's activation as
\begin{equation}
\label{eqn:act_quant}
\begin{gathered}
{y^l} = PACT\left( {{x^l}} \right) = \frac{{\left| {{x^l}} \right| - \left| {{x^l} - \left| {{\alpha ^l}} \right|} \right| + \left| {{\alpha ^l}} \right|}}{2},\\
{y^l} = {Q_k}\left( {{y^l}/\left| {{\alpha ^l}} \right|} \right) \cdot \left| {{\alpha ^l}} \right|.
\end{gathered}
\end{equation}
$x^l$ is the activation of layer-$l$. $PACT(\cdot)$ is a function that clips the activation $x^l$ to the range between $[0, {\left| {{\alpha ^l}} \right|}]$. $\alpha^l$ is a layer-wise trainable upper bound, determined by the training of the network. It is observed that during training $\alpha ^l$ can sometimes become a negative value, which affects the correctness of the PACT \cite{choi2018pact} function. To ensure $\alpha ^l$ is always positive and to increase training stability, we use the absolute value of the trainable parameter $\alpha ^l$ rather than its original value. $y^l$ is the clipped activation from layer-$l$ and it is further quantized to $y_k^l$, a k-bit activation tensor. Note that activations from the same layer share the same floating point coefficient $\alpha^l$, but activations from different layers can have different coefficients. This is problematic for the concatenative skip connection, since if the coefficients $\alpha^l$ and $\alpha^{l-1}$ are different, we need to first cast $y_k^{l-1}$ and $y_k^{l}$ from fixed-point to floating point, re-calculate a coefficient for the merged activation, and quantize it again to new fixed-point numbers. This process is very inefficient. 

In our experiment, we notice that most of the layers in the DiracDeltaNet have similar coefficients with values. Therefore, we rewrite equation (\ref{eqn:act_quant}) as 
\begin{equation}
\label{eqn:act_quant_new}
{y^l} = {Q_k}\left( {{y^l}/\left| {{\alpha ^l}} \right|} \right) \cdot {\left| {s} \right|}.
\end{equation}
where $s$ is a coefficient shared by the entire network. This step ensures that activations from different layers of the network are quantized and normalized to the same scale of $[0, {\left| {s} \right|}]$. As a result, we can concatenate activations from different layers directly without extra computation. Moreover, by using the same coefficient $s$ across the entire network, the convolution can be computed completely via fixed-point operations. The coefficient $s$ can be fixed before or leave it as trainable. A general rule is that we should let $s$ have similar values of $\alpha^l$ from different layers. Otherwise, if $s/\alpha^l$ is either too small or too large, it can cause gradient vanishing or exploding problems in training, which leads to a worse accuracy of the network.

In our network, we merge the PACT function and activation quantization into one module and name it ActQuant. The input to ActQuant is the output of 1$\times$1 convolutions. Since the input and weight of the convolution are both quantized into fixed-point integers, the output is also integers. Then, ActQuant is implemented as a look-up-table whose parameters are determined during training and fixed during inference.

\begin{figure}[!t]
\begin{center}
\centering 
\includegraphics[width=0.9\linewidth]{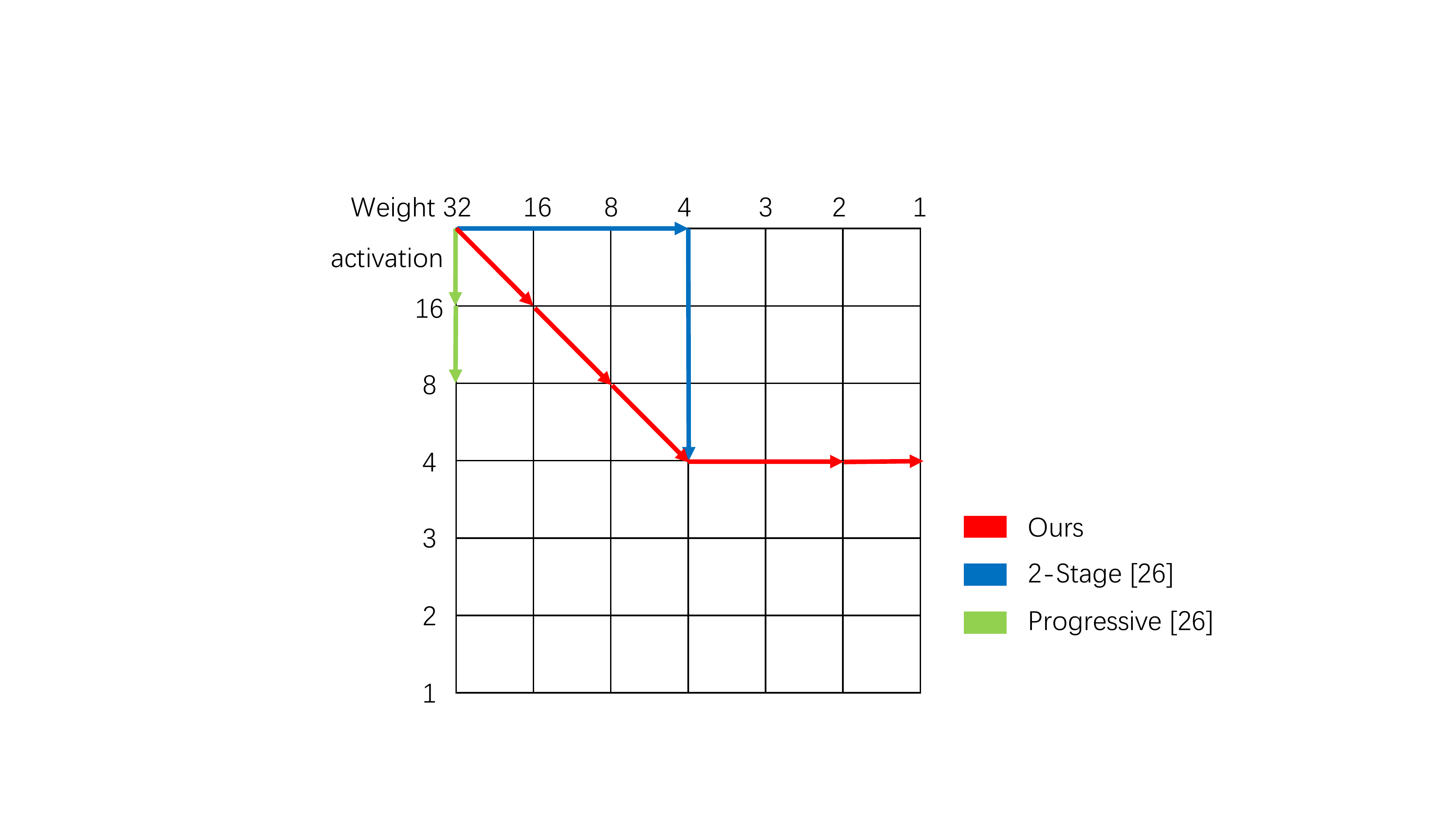}
\vspace{-13pt}
\caption{Quantization Grid}
\label{fig:grid}
\end{center}
\vspace{-5pt}
\end{figure}

We follow \cite{Zhuang2017progressive} to quantize the network progressively from full-precision to the desired low-precision numbers. The process is illustrated in Fig. \ref{fig:grid}, where x-axis denotes bit-width of weights and y-axis denotes the bit-width of activations. We start from the full-precision network, train the network to convergence, and follow a path to progressively reduce the precision for weights or activations. At each point, we fine-tune the network for 50 epochs with step learning rate decay. Formally, we denote each point in the grid as a quantization configuration ${\mathscr{C}_{w,a}}\left( N_{w} \right)$. Here $w$ represents the bitwidth of weight. $a$ is the bitwidth of activation. $N_w$ is the network containing the quantized parameters. The starting configuration would be the full precision network ${\mathscr{C}_{32,32}}\left( N_{32} \right)$. Starting from this configuration, one can either go down to quantize the activation or go right to reduce the bitwidth of weight. More aggressive steps can be taken diagonally or even across several grids. The two-stage and progressive optimization methods proposed in \cite{Zhuang2017progressive} can be represented as two paths in Fig. \ref{fig:grid}. 

In our work, we start from ${\mathscr{C}_{32,32}}\left( N_{32} \right)$. Then we use $N_{32}$ to initialize $N_{16}$ and obtain ${\mathscr{C}_{16,16}}\left( N_{16} \right)$. And we apply step lr decay fine-tuning onto $N_{16}$ to recover the accuracy loss due to the quantization. After several epochs of fine-tuning, we get the desired low-precision configuration ${\mathscr{C}_{16,16}}\left( N_{16}' \right)$ with no accuracy loss. Following the same procedures, we are able to first go diagonally in the quantization grid to ${\mathscr{C}_{4,4}}\left( N_{4} \right)$ with less than 1\% top-5 accuracy loss compared to its full precision counterpart.

\begin{table}
\caption{Quantization Result on DiracDeltaNet}
\vspace{-13pt}
\centering
\begin{tabular}{c|c|c}
 & full   & w4a4  \\ 
\hline
Top-1 Acc & 68.9\% & 68.3\%  \\
Top-5 Acc & 88.7\% & 88.1\% 
\end{tabular}
\label{tab:quantization_result}
\vspace{-15pt}
\end{table}

We use a pre-trained ResNet50 label-refinery \cite{bagherinezhad2018label} to boost the accuracy of the quantized model. Even with such low-precision quantization, our quantized model still preserves a very competitive top-5 accuracy of 88.1\%. Most of the previous quantization works \cite{choi2018pact, Zhuang2017progressive, zhou2016dorefa} are only effective on large models such as VGG16, AlexNet or ResNet50. 
Our quantization result is summarized in Table \ref{tab:quantization_result}.

\section{Hardware Design}
As mentioned in section \ref{DiracDeltaNet}, we aggressively simplified ShuffleNetV2's operator set. Our modified network is mainly composed of the following operators:
\begin{itemize}
    \item $1\times1$ convolution
    \item $2\times2$ max-pooling
    \item shift
    \item shuffle and concatenation
\end{itemize}
Our accelerator, Synetgy, is tailored to only support the operators above. 
This allows us to design more specialized compute units with simpler control, which enables us to further improve the hardware efficiency. 
The compute of the fully-connected layer can be mapped onto our convolution unit. 
Shuffle operation is not fully supported on FPGA. CPU-based memory copy is needed to maintain the memory layout.
And the remaining average-pooling layer which is not supported on the FPGA is offloaded to the ARM processor on the SoC platform. 

The benefits of simplified operator come from the algorithm-hardware co-design, which also increase the productivity of hardware implementation. The accelerator implementation only took two people working for one month using HLS.

\subsection{The accelerator architecture}

\begin{figure}[H]
\centering
\vspace{-5pt}
\includegraphics[width=1\linewidth]{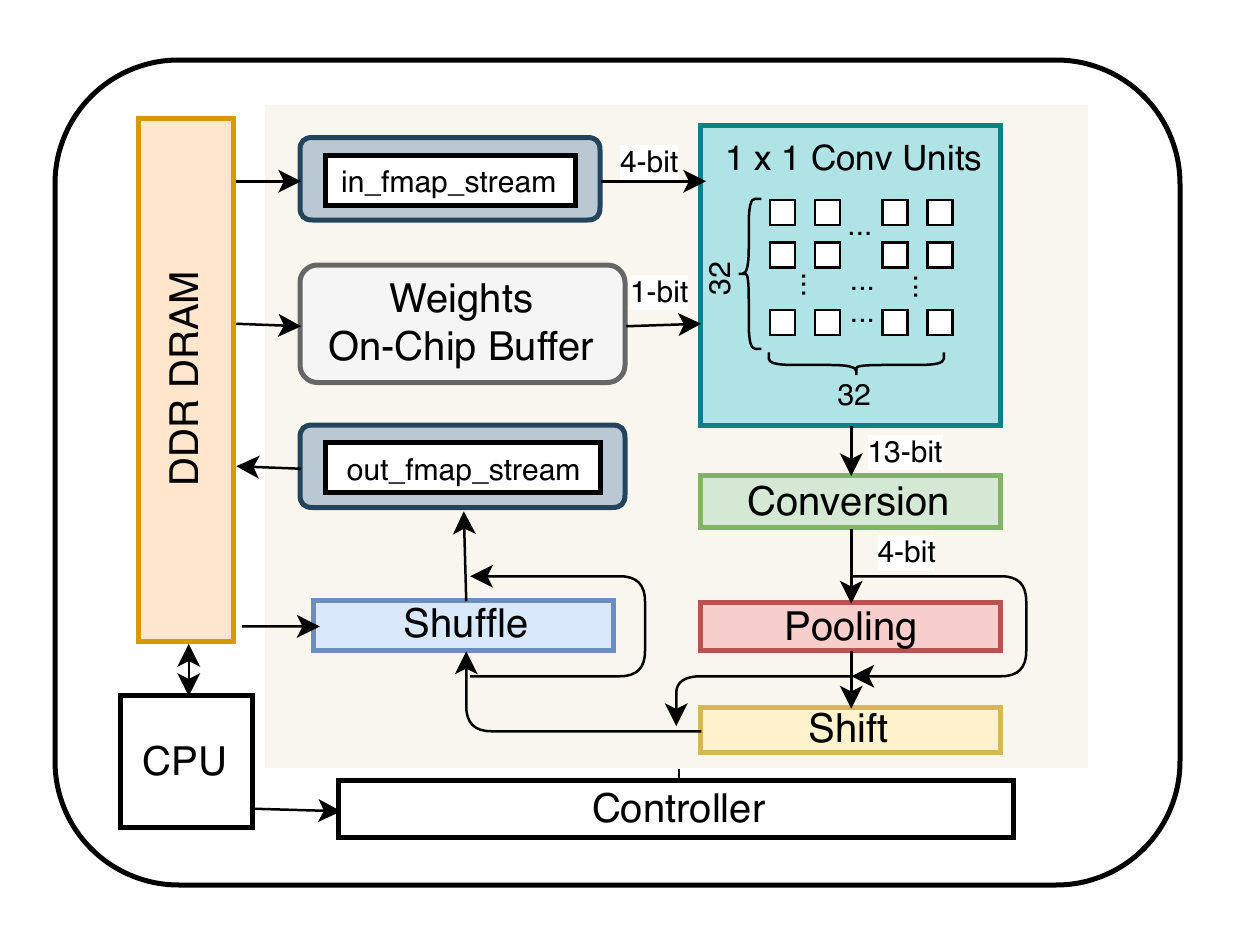}
\vspace{-20pt}
\caption{Accelerator Architecture}
\label{fig:accel_arch}
\vspace{-5pt}
\end{figure}

Fig. \ref{fig:accel_arch} shows the overall accelerator architecture design. Our accelerator, highlighted in light yellow, can be invoked by the CPU for computing one $1\times1$ Conv-Pooling-Shift-Shuffle subgraph at a time. The CPU provides supplementary support to the accelerator. Both the FPGA and the CPU are used to run the network. 
In quantized DiracDeltaNet, weights are 4-bit, input and output activations are 4-bit, and the largest partial sum is 17-bit. The width of partial sum is determined by the input feature bit width and the largest channel size. Given that the largest channel size is 512, there are $2^4\times2^4\times512$ possible outcomes from the convolution, which requires 17 bits to represent.
\begin{table}[]
\caption{Notations}
\vspace{-5pt}
\resizebox{0.45\textwidth}{!}{%
\begin{tabular}{l|l|l}
\hline
Notation & Type         & Description                             \\ \hline
WIDTH    & variable     & width of feature map                    \\
HEIGHT   & variable     & height of feature map                   \\
IC\_TOTAL & variable     & total input channel size                \\
OC\_TOTAL & variable     & total output channel size               \\
IC   & constant: 32  & parallelism on input channel dimension  \\
OC       & constant: 32 & parallelism on output channel dimension \\
\hline 
\end{tabular}%
}
\label{tab:notation}
\vspace{-10pt}
\end{table}
\subsubsection{Dataflow Architecture}
Our hardware design is based on the dataflow architecture template \cite{cheng2016high, vivado2018ug}.
As illustrated in Fig.~\ref{fig:accel_arch}, we first extract a few process functions from the major operations including $1\times1$ convolution, $2\times2$ max-pooling, shift, shuffle and the memory load and store. 
We then chain them together using FIFOs with blocking read and non-blocking write. Note that the write is blocking once the FIFO is full. 
All the process functions are running concurrently and the execution of each function is triggered by the arrival of the data. 
Therefore, more task-level parallelism can be explicitly exposed to the HLS tool in addition to the instruction-level parallelism.

\subsubsection{Convolution Unit}
\begin{figure}[H]
\centering
\includegraphics[width=3in]{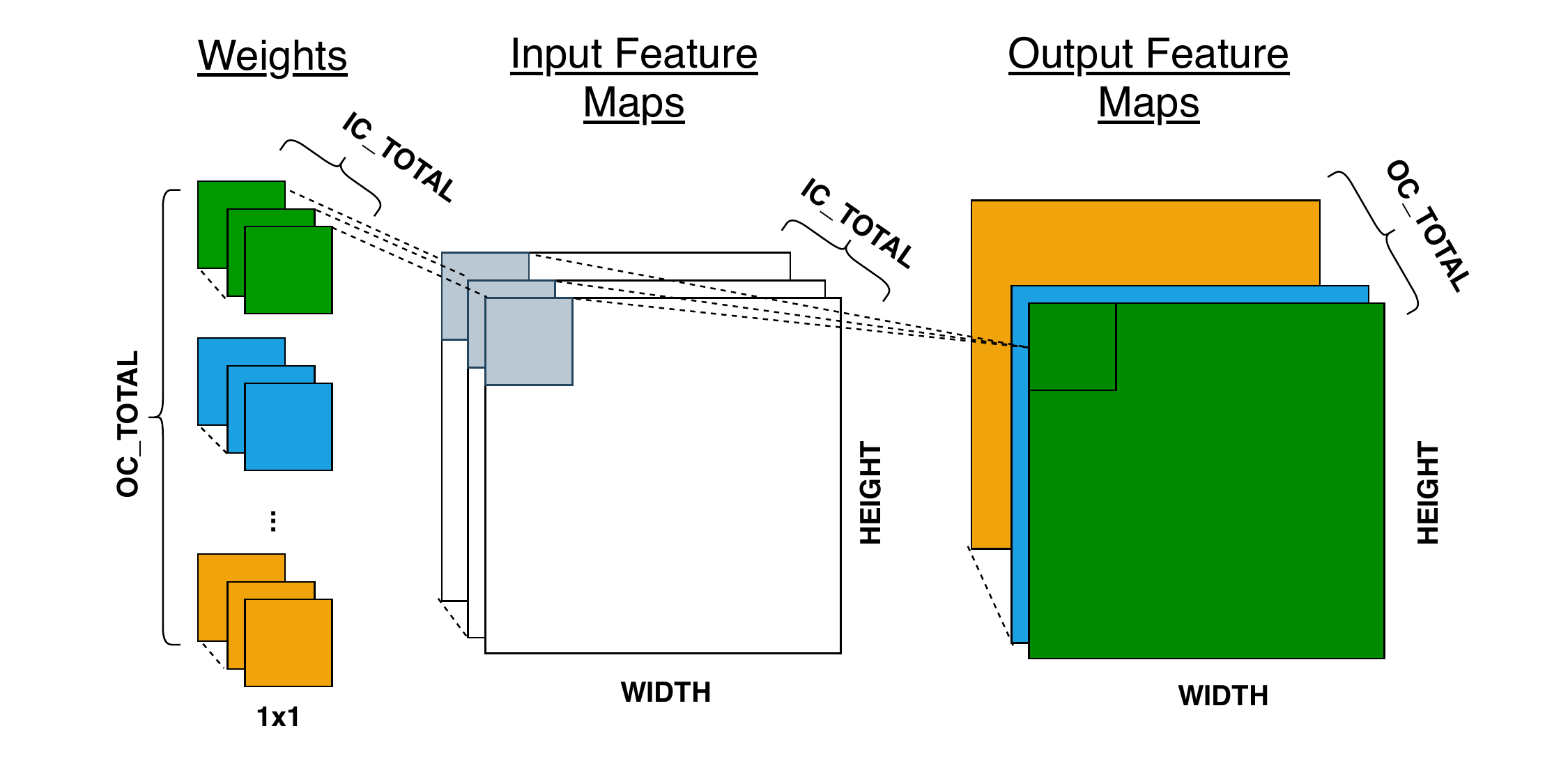}
\caption{1$\times$1 Convolution}
\label{fig:1x1conv}
\end{figure}

The notations used in this section are listed in Table \ref{tab:notation}. As shown in Fig. \ref{fig:1x1conv}, given an input feature map of size $ WIDTH \times HEIGHT \times IC\_TOTAL$ and a weight kernel of size $IC\_TOTAL \times OC\_TOTAL$, the generated output feature map is of size $WIDTH \times HEIGHT \times OC\_TOTAL$ in 1$\times$1 convolution. The  1$\times$1 convolution is essentially a matrix-matrix multiplication.  
 
\vspace{-13pt}

Although \cite{kwon2018co} suggests a weight stationary dataflow for 1 $\times$ 1 convolution dominant ConvNets, we find it not applicable to our design as the bit width of weights is much smaller than the partial sums (4 bit vs 17 bits). Transferring the partial sums on and off-chip will incur more traffic on the memory bus. Therefore, we adopt the output stationary dataflow by retaining the partial sums in the local register file until an output feature is produced. 
 
\begin{figure}[h]
\centering
\frame{
\includegraphics[width=0.42\textwidth]{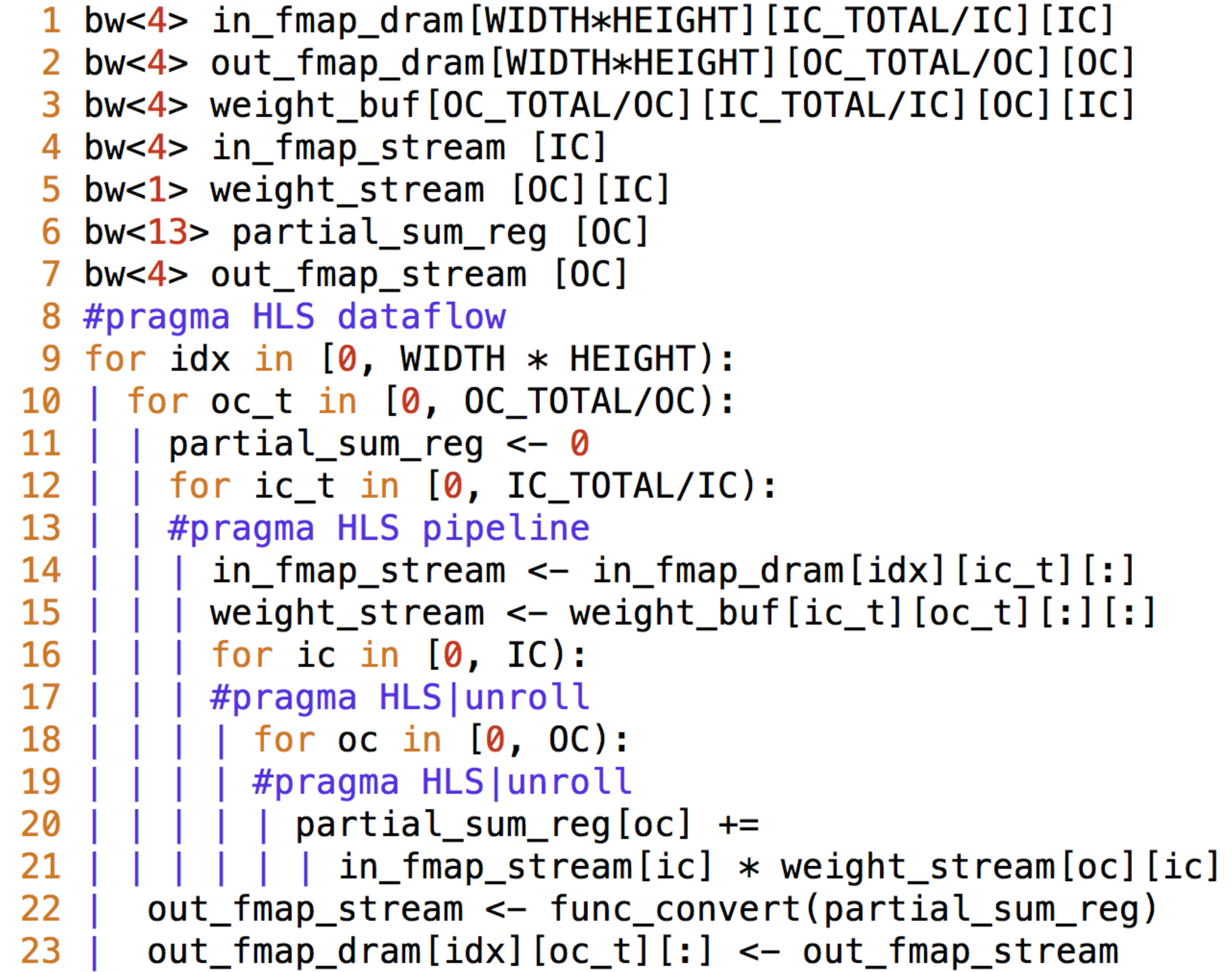}}
\caption{Pseudo Code for Kernel Compute Scheduling}
\label{fig:code_snippet}
\vspace{-15pt}
\end{figure} 
 
Fig. \ref{fig:code_snippet} shows how we schedule the workload onto the accelerator. Note that the nested loops starting at line 17, 19 are automatically unrolled.
Weights are prefetched onto on-chip BRAM $weight\_buf$. 
We first block our inputs so $IC \times OC$ multiplications can be mapped onto the compute units at each iteration (Line 13$\sim$21). In every iteration, $IC$ input features are fetched from the DRAM.
They are convolved with $OC$ number of weights of size $IC$ and produce $OC$ partial sums. 
Each iteration of the loop nest along the input channel dimension at line 12 takes $7\sim38$ cycles to finish based on the Vivado HLS report. 
Equivalently, it takes $7 \sim 38$ cycles to finish $IC \times OC$ 4/4 bit multiplication. 
The partial sums are stored in the registers, which can be simultaneously accessed in every cycle. 
The parameter $IC$ and $OC$ were tuned for the area performance tradeoff. Increasing them increases overall resource utilization but helps to reduce the total number of execution cycles.

Based on the roofline model~\cite{williams2009roofline}, the attainable throughput is the compute-to-communication (CTC) ratio multiplied by the bandwidth when it is bandwidth bound. The CTC ratio of our compute unit for the input feature is $OC\_TOTAL$ (maximum number is 512 in DiracDeltaNet), which is a variable. 
Larger output channel size indicates higher CTC ratio. 
According to our measurement, the maximum bandwidth of the DDR channel is 6GB/s, which means $6 \times 2$ Giga input features (1 Byte contains two 4-bit features) can be loaded. 
The theoretical memory bound throughput should be $512 \times 6 \times 2 = 6144$GMACs $=12288$GOPs. For compute bound problems, the attainable throughput is dependent on the compute capability. In our case, it is $IC \times OC \times freq = 32 \times 32 \times 250MHz = 256GMACs $=$512$GOPs. Based on the analysis, the convolution unit will reach the bandwidth bound before it hits the computation roofline.

\begin{figure}[h]
\centering
\includegraphics[width=0.45\textwidth]{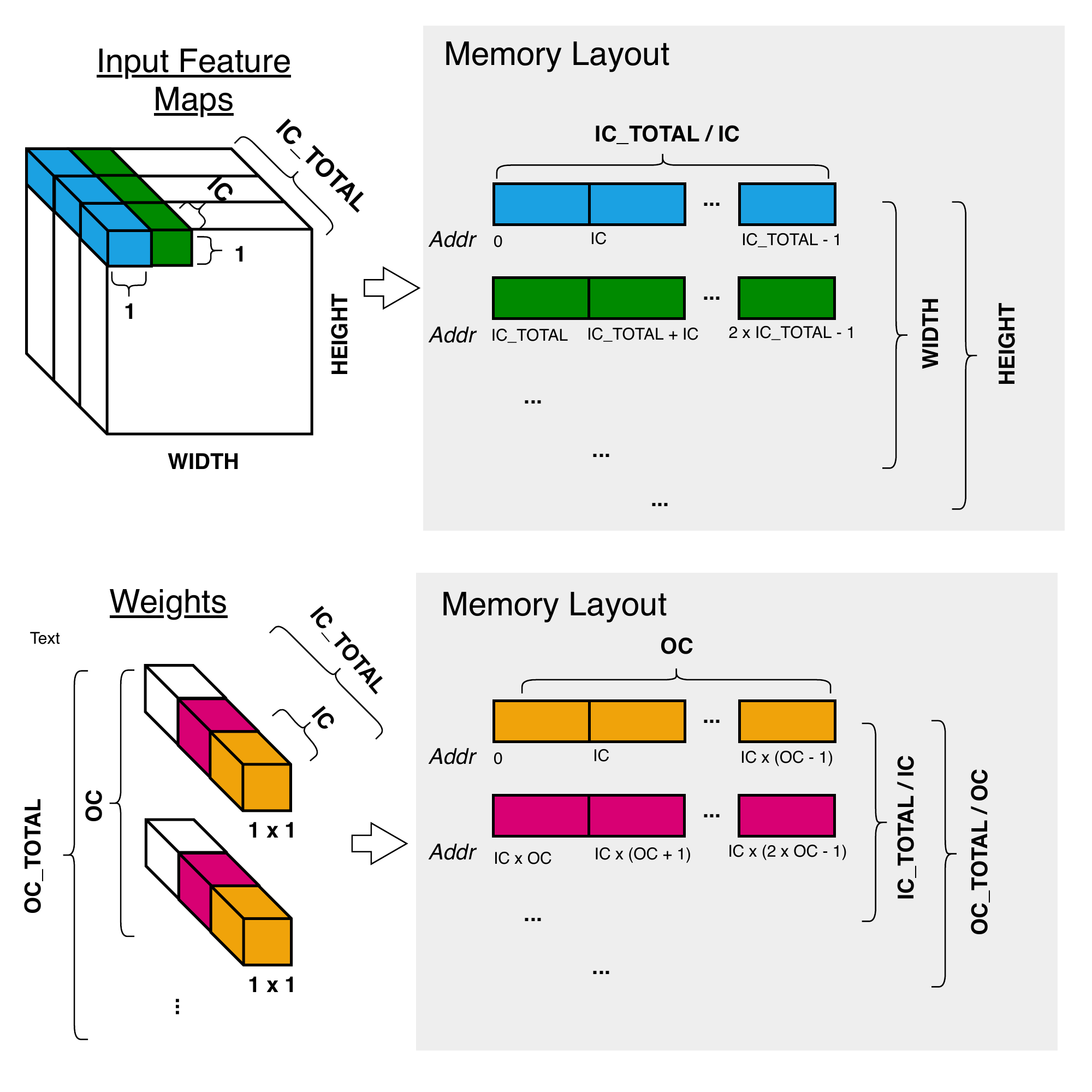}
\vspace{-10pt}
\caption{Input Layout in DRAM}
\label{fig:input_layout}
\vspace{-10pt}
\end{figure} 

\subsubsection{Conversion Unit}
\label{sec:conversion}
The high bitwidth to low bitwidth conversion is performed immediately after the kernel computation. It is a step function with 16 intervals that converts 17-bit partial sum to 4-bit activation. The threshold values are different for each layer. 
All of the read-only threshold values are stored in on-chip BRAMs. 
An index number should be specified by the user function to select which set of threshold values to use for the compute of the current layer. 
In hardware, this unit is implemented by using 16 comparators. They are mapped onto a binary tree structure to reduce the circuit latency. 

\subsubsection{Pooling Unit}
We adopt the line buffer design described in \cite{zhao2017accelerating} to implement the $2\times2$ max-pooling layer. 
For every iteration, $(WIDTH + 1)$ of $IC$ deep pixels are first fetched into the line buffers.  Once the next pixel value is fetched, a $2\times2$ large sliding window is formed. For every 2 cycles, we compare the values in the $2\times2$ sliding window, output the largest one, and fetch the next 2 values. It takes $IC\_TOTAL/IC$ iterations to finish the compute.

\subsubsection{Shift Unit}
The line buffer design is also used for the shift operation. 
In the shift unit, the input images are first padded with 1 zero-value pixel at the width and height dimension. 
$(2\times (WIDTH+2) + 2)$ of pixels are then buffered and a $3\times3$ sliding window is formed.
The shift direction is different for different input channels. It is calculated based on the input channel index. 
After initialization, the unit is able to produce $1$ output pixel per cycle.

\subsubsection{Shuffle Unit}
Shuffle is implemented by changing the address offset of output features during the writeback phase. Since the shuffle operation still requires us to concatenate the outputs from the previous DiracDeltaNet block to the current DiracDeltaNet block outputs, the CPU is used to copy the output from previous DiracDeltaNet unit to the shuffled address. The memory copy operation should be done concurrently with the computation of current DiracDeltaNet unit. 

\subsubsection{Fully Connected Unit}
We don't explicitly design a dedicated unit to compute FC layer. Instead, we map the compute of FC layer onto our existing hardware convolution unit. The feature map size is 1 for the FC layer. While the convolution unit only supports 4-bit weight, the FC layer's computation is mapped in a bit serial like manner. The convolution unit processes each bit of the FC weight iteratively and bit shift is done by configuring the step function in the conversion unit.

\subsection{Software}
We use the ARM processor to control the layer-based accelerator and to compute the last $7\times7$ average-pooling layer that is not supported by the accelerator.  
The host application runs on a full Linux system on the ARM CPU, which controls the memory-mapped accelerator through the UIO driver interface. The Xilinx python-based PYNQ APIs \cite{xilinx2018pynq} are used for fast deployment of the host software code on the Ultra 96 board.

\subsection{Experimental Results}

We implement our accelerator, Synetgy, on the Ultra96 development board with Xilinx Zynq UltraScale+ MPSoC targeted at embedded applications. Table \ref{tab:area} shows the overall resource utilization of our implementation. We are able to utilize 73\% of the total LUTs on the FPGA, as the bit-level 4/4bit multiplications are mapped onto LUTs.
BRAMs are mainly used for implementing the FIFO channels.  
DSPs are used for the address calculation for the AXI protocol. 
Our implementation runs at 250 MHz. Power measurements are obtained via a power monitor. We measured 5.3W with no workload running on the programming logic side and 5.5W max power on the Ultra96
power supply line when running our network.

\begin{table}[htbp]
\vspace{-5pt}
\caption{Resource Usage}
\vspace{-5pt}
\centering
\begin{tabular}{|c|c|c|c|c|}
\hline
\textbf{LUT} &\textbf{FF} & \textbf{BRAM} & \textbf{DSP} \\
\hline
51776 (73.4\%) & 42257 (29.9\%) & 159 (73.6\%) & 360 (100\%) \\ \hline
\end{tabular} 
\label{tab:area}
\end{table}

\begin{table*}[ht]
\vspace{-15pt}
\caption{Performance Comparison of Synetgy and Previous Works}
\vspace{-5pt}
\begin{tabular}{|l|l|l|l|l|l|l|l|}
\hline
\textbf{}         & \textbf{VGG-SVD\cite{qiu2016going}} & \textbf{AlexNet\cite{liang2018fp}} & \textbf{VGG16\cite{suda2016throughput}} & \textbf{VGG16 \cite{guo2017software}} & \textbf{DoReFa\cite{jiao2017accelerating}} & \textbf{FINN-R \cite{blott2018finnr}} & \textbf{Ours}   \\ \hline
Platform  & Zynq XC7Z045   & Stratix-V  & Stratix-V    & Zynq 7Z020   & Zynq 7Z020   & Zynq ZU3EG  & Zynq ZU3EG   \\ \hline
Frame Rate (fps)  & 4.5  & \bf{864.7}  & 3.8   & 5.7  & 106.0  & 200.0  & 66.3\\ \hline
Top-1 Acc & 64.64\%   & 42.90\%  & 66.58\%  & 67.72\%  & 46.10\%  & 50.3\%  & \bf{68.30\%}  \\ \hline
Top-5 Acc  & 86.66\%  & 66.80\%  & 87.48\%  & 88.06\%   & 73.10\%  & N/A  & \bf{88.12\%} \\ \hline
Precision & 16b  & 16b  & 8-16b  & 8b  & 2b  & 1-2b  & 4-4b \\ \hline
Frequency(MHz)  & 150 & 150  & 120  & 214  & 200 & 220  & 250  \\ \hline
Power(W)  & 3.0 & 26.2 & 19.1  & 3.0  & 2.3 & 10.2  & 5.5 \\ \hline
\end{tabular}
\label{tab:time}
\vspace{-5pt}
\end{table*}

\begin{table}[]
\caption{Frame Rate on Different Batch Size}
\vspace{-5pt}
\begin{tabular}{|c|cccccc|}
\hline 
Batch Size  & 1 & 2 & 4 & 8 & 16 & \\ \hline
Frame Rate (fps)  & 41.4  & 53.6 & 62.6 & 65.6 &  66.3& \\    
\hline
\end{tabular}
\label{tab:batchsize}
\vspace{-10pt}
\end{table}

We compare our accelerator against previous work in Table~\ref{tab:time}. As explained before, ConvNets for ImageNet classification are usually orders of magnitude more complex than CIFAR10 classification. Therefore, we only compare accelerators targeting ConvNets for ImageNet classification with reasonable accuracy. Our work focuses on achieving competitive accuracy while improving the actual inference speed in terms of frames per second. Our experiments show that we successfully achieve those two goals. From the table, we can make the following observations: 
1) Synetgy achieves the highest top-1 and top-5 accuracy on ImageNet. The only previous work that comes close to our accuracy is \cite{guo2017software}, but its frame rate is 11.6$\times$ slower than ours.  
2) Among the embedded accelerators whose top-1 accuracy is higher than 60\%, which is a loose constraint, our model achieves the fastest inference speed. 
3) Without the accuracy constraint, the speed of \cite{liang2018fp, jiao2017accelerating, blott2018finnr} can go as fast as 864.7 frames per second. But their accuracy is rather low.
4) The peak attainable throughput of our accelerator is 418 GOPs, which is close to the theoretical compute roofline.  
Our average throughput (47.09 GOPs) is currently limited by the low hardware utilization. The inefficiency is mainly from the software shuffle operations and the first convolution layer whose input dimension is 3 which is much less than the hardware tiling factor $IC$.
However, Synetgy still achieves competitive frame rate, demonstrating the efficacy of our co-design methodology.   
We see the opportunity of significant frame rate improvement through further algorithm-hardware co-design.

The reported frame rate is achieved with batch size set to 16. There is a fixed software overhead for invoking the poll-based hardware accelerator. 
The computation latency of the DiracDelta Block1 in Table~\ref{tab:block} is 0.15ms when the batch size is equal to 1. The latency for a single read on the accelerator control register is 0.40ms, which is greater than the actual compute time. 
In order to minimize this software overhead, we increase the batch size to schedule more computation running on the accelerator per invocation. 
Furthermore, the weights stored in on-chip BRAM get reused more when batch size is increased. 
The frame rates of implementations with different batch sizes are summarized in Table~\ref{tab:batchsize}.

We break down the runtime of the whole heterogeneous system by bypassing one component of the system and measure the runtime.  
We observe that the CPU-based memory copy for the shuffle operation significantly degrades the performance. 
However, all other non-conv components (sw average pooling, FC, PYNQ API call) impact the overall performance slightly.

To further understand the efficiency of various operators (1$\times$1 conv, 2$\times$2 max-pooling, shift, and shuffle) implemented on FPGA and CPU, we measure the runtime of the DiracDeltaNet blocks with different configurations on Synetgy. The result is summarized in Table~\ref{tab:block}. We test 2 blocks with different input feature map and channel sizes. Note that the theoretical OPs of Block1 and Block2 is the same. As shown in the table, pooling and shift incur almost no performance drop. 
This is because the process functions for performing these operations do not impose new bottlenecks on the dataflow pipeline. 
Software memory copy latency of shuffle is more significant on Block1 than Block2. This is because memory copy overhead is proportional to $HEIGHT\times WIDTH \times OC\_TOTAL$. But total OPs $HEIGHT \times WIDTH \times IC\_TOTAL \times OC\_TOTAL$ remains the same, which means that smaller feature map needs less time for memory copy. The memory copy overhead can be possibly alleviated through running bare-metal C code on the CPU. 

\begin{table}
\centering
\caption{Runtime Analysis for the First and Last DiracDeltaNet Blocks in Different Operator Configurations (Batch=10)}
\vspace{-5pt}
\begin{tabular}{|c|c|c|}
\hline
                 & \multicolumn{2}{c|}{Runtime (ms)}                 \\ 
\hline
                 & \multicolumn{1}{c|}{\textbf{Block1}} & \multicolumn{1}{c|}{\textbf{Block2}}  \\ 
\hline
feature map size & 28                          & 7                           \\
in\&out channel  & 128                         & 512                         \\ 
\hline
conv only       & 1.531                       & 0.989                       \\
conv+pool        & 1.530                      & 0.993
\\
conv+shift       & 1.537                      & 0.996                       \\
conv+shuffle     & 4.409                       & 1.636                       \\
overall          & 4.364                       & 1.441     
\\
\hline
\end{tabular}
\vspace{-17pt}
\label{tab:block}
\end{table}

\section{Conclusion and Future Works}
In this paper, we adopt an algorithm-hardware co-design approach to develop a ConvNet accelerator called Synetgy and a novel ConvNet model called DiracDeltaNet. Based on ShuffleNetV2, we optimize the network's operators by replacing all the 3$\times$3 convolutions with shift operations and 1$\times$1 convolutions. This allows us to build a compute unit exclusively customized for 1$\times$1 convolutions for better efficiency. We quantize the network's weights to 4-bit and activations to 4-bit fixed-point numbers with less than 1\% accuracy loss. These quantizations very well exploit the nature of FPGA hardware. As a result, DiracDeltaNet has a small parameter size, low computational OPs, hardware-friendly skip connections, low precision, and simplified operators. These features allow us to implement highly customized and efficient accelerators on FPGA. We implement the network on Ultra96 Soc systems. The implementation only took two people one month using HLS tools. Our accelerator, Synetgy, achieves a top-5 accuracy of 88.1\% on ImageNet, the highest among all the previously published embedded FPGA accelerators. It also reaches an inference speed of 66.3 FPS, surpassing prior works with similar accuracy by 11.6$\times$. While we see many more opportunities for further optimization, we believe this demonstrates the efficacy of our co-design methodology. 

For the future works, we will focus on further optimization. 
For example, we can add more layers in the dataflow architecture to improve the compute-to-communication ratio. 
Correspondingly, we will need to adjust the network such that the computation subgraphs are more symmetric.

\begin{acks}
We would like to thank all of the people who helped us realize this project, especially the anonymous reviewers, Kostadin Ilov, Rock Qu, Alessandro Pappalardo, Amir Gholaminejad, Peter Jin, Ravi Krishna, and Alvin Wan. 
The information, data, or work presented herein was funded in part by the Advanced Research Projects Agency-Energy (ARPA-E), U.S. Department of Energy, under Award Number DE-AR0000849. The Research was partially funded by ADEPT Lab industrial sponsor Intel, and ADEPT Lab affiliates Google, Siemens, and SK Hynix. The views and opinions of authors expressed herein do not necessarily state or reflect those of the United States Government or any agency thereof.

\end{acks}

\bibliographystyle{unsrt}
\bibliography{ref}

\end{document}